\documentclass{article}

\usepackage[numbers]{natbib}
\usepackage{graphicx}

\usepackage[final]{neurips_2022}

\usepackage[utf8]{inputenc} 
\usepackage[T1]{fontenc}    
\usepackage{hyperref}       
\usepackage{url}            
\usepackage{booktabs}       
\usepackage{amsfonts}       
\usepackage{nicefrac}       
\usepackage{microtype}      
\usepackage{xcolor}         
\usepackage{graphicx}
\usepackage{longtable}
\usepackage{url}
\usepackage[export]{adjustbox} 

\title{Continual learning on deployment pipelines for Machine Learning Systems}

\author{
Qiang Li\thanks{Thanks to the side-project support from Accenture CIE IT/OT Group (about the author https://qiangli.de)
    } \\
  Accenture \\
  Kaistraße 20, 40221 Düsseldorf \\
  \texttt{qiang.i.li@accenture.com} \\
  \And
  Chongyu Zhang \\
  Technical University of Munich \\
  Arcisstraße 21, 80333 München\\
  chongyu.zhang@tum.de \\
}

\begin{document}

\maketitle

\begin{abstract}
  
Following the development of digitization, a growing number of large Original Equipment Manufacturers (OEMs) are adapting computer vision or natural language processing in a wide range of applications such as anomaly detection and quality inspection in plants. Deployment of such system is becoming a extremely important topics. Our work starts with the least-automated deployment technologies of machine learning systems, includes several iterations of updates, and ends with a comparison of automated deployment techniques. The objective is, on the one hand, to compare the advantages and disadvantages of various technologies in theory and practice, so as to facilitate later adopters to avoid making the generalised mistakes when implementing actual use-cases, and thereby choose a better strategy for their own enterprises. On the other hand, to raise the awareness of the evaluation framework for the deployment of machine learning systems, to have more comprehensive and useful evaluation metrics (e.g. table \ref{comparsion} and figure 2 \ref{evaluationmetric}), rather than only focusing on a single factor (e.g. company's cost). This is especially important for decision-makers in the industry.
  
\end{abstract}

\section{Introduction}

Increasingly, large-scale machine learning systems including front and back end, cloud platforms, and integrated ecosystems of computer vision or natural language processing are being deployed and leveraged with great value. Meanwhile, a variety of state-of-the-art models and the pursuit of innovative techniques in academia are in turn facilitating this digital revolution. However, we can note that the discussions on a specific technology are far more heated than the discussions on the need to use that technology \cite{datamining2}, and it is unfortunate that with the trend of KPIs in the industry, we tend to miss the fairness of evaluating deployment technologies from a more diverse perspective. 

It is worth noting that all the experiments and conclusions in this paper have been tested in industry-specific cases, and this paper can be used both as a technical handbook for the deployment of ML System (e.g. quality inspection systems), also as a valuable reference for the selection of different technologies.

\section{Methodology}

\subsection{How Edge Device and NSSM Batch Services enable Static Deployment}

For most industrial applications, it is reasonable to build an MVP demo and implement some well-known algorithms for quality inspection systems, such as ResNet \cite{resnet}  or Fast-RCNN \cite{FastRCNN}, based on well-established TensorFlow Object Detection API \cite{ObjectAPI}. Without the support of the cloud platform, through NSSM \cite{NSSM} and Edge device, it is also possible to quickly achieve the deployment and construction. You may doubt the stability of those systems. On the contrary,  it is often possible to achieve a real industrial use-case of recognition with a significant amount of data volume (e.g., 3 T image data per day) through these tools. Figure 3  and Figure 4 describe the specific implementation of NSSM Batch Service-based Deployment in great detail.

Although it is stable and appropriate for the initial stage of the product, this deployment technology also reveals some weaknesses, such as the complex deployment steps that often bother some senior AI engineers. The second is very poor scalability, and if the model goes wrong, this often does not achieve automatic deployment, demanding killing or suspending the previous batch service. The port connectivity issues, how to use it, and which port to use require much attention. In addition, due to the existence of edge devices, we often need a third-party virtual machine connection tool like NetSupport Manager (see Figure 3). When deploying large-scale machine learning systems, with the increasing complexity of use cases and the cooperation of the various regions, this deployment method of installing a separate edge device and NSSM batch service is often not recommended \cite{NSSM}.

\subsection{Semi-Automatic Deployment using Jenkins and OpenShift}

The invention of Jenkins \cite{Jenkins} makes the deployment of models much easier, and of course, Jenkins is not the only technology utilized here. The above Figure 1 nicely illustrates the entire deployment process from docker generation, uploading to the Openshift container storage, and finally, the trigger and scaling process.

\begin{figure}[t]
  \centering
  \label{Jenkinspipline}
  \includegraphics[width=\columnwidth]{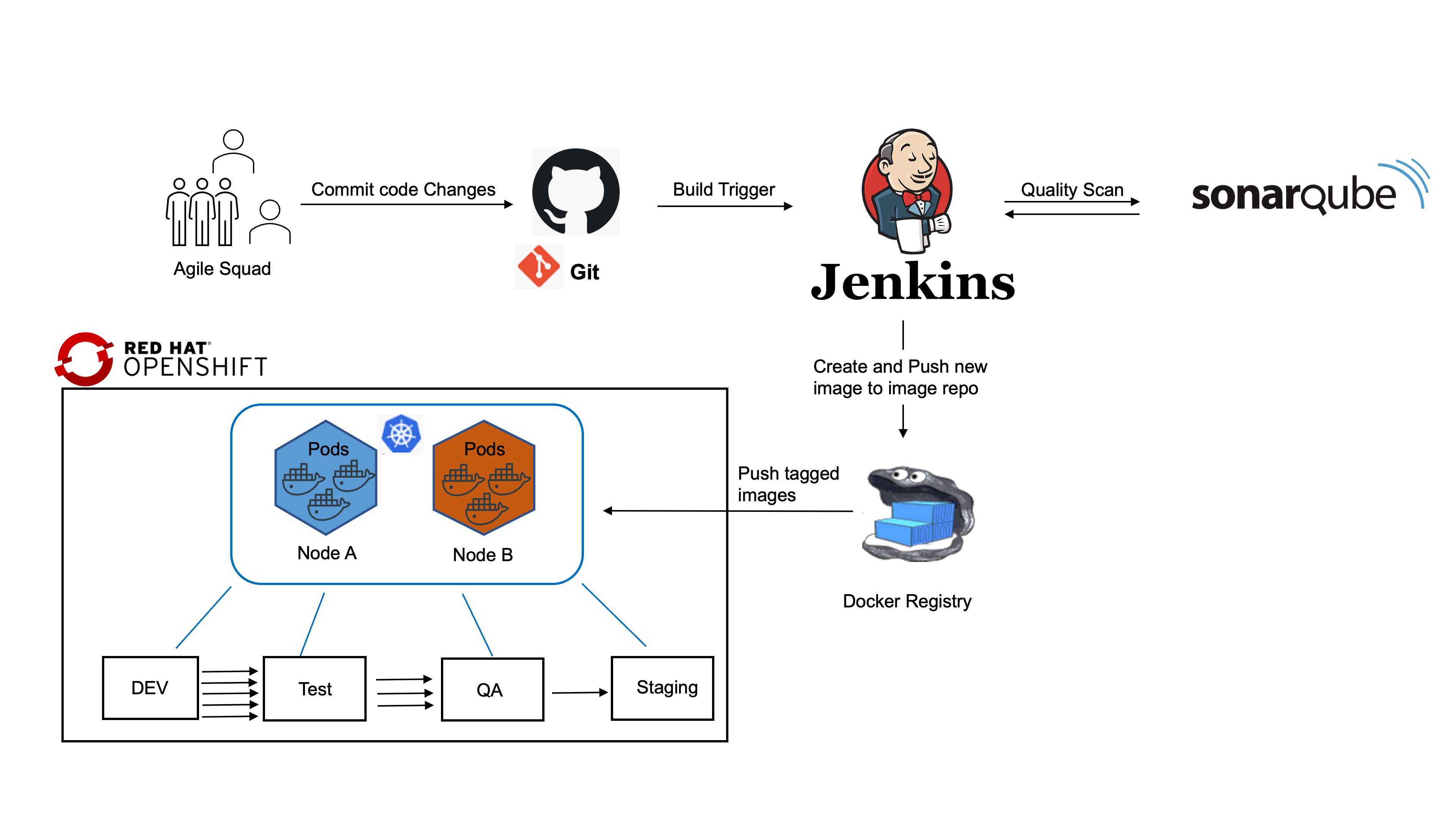} 

  \caption{Jenkins and OpenShift CI/CD Pipeline \cite{Jenkins}. First upload the model to the cloud (e.g., model directory), then update the business logic code by git; Jenkins acts here as a monitor and can be triggered to build the docker image. Then the docker image name and the tag will be forwarded to the OS \cite{openshiftonline}, and the AI container will be saved in Nexus. Finally, the OS can use the latest docker image build to complete the deployment. Only the settings file about the application needs to be set up here, where we can very easily define the speed of sending images, the processing time for each inference, and the CPU/resources for each use case.}
\end{figure}

As shown in Figure 6, many Jenkins operations are graphical, making the deployment process more intuitive. The user only needs to click on some build buttons to trigger docker image builds. However, this deployment method also has a significant drawback, namely the accessibility of the logs. In some cases, it is unclear which part of the deployment process has failed due to the limited length of the log. Furthermore, it is deplorable that the docker image is directly connected to OpenShift, so getting the log and the docker image for a local test is not very straightforward. It is frequently only by the end of the deployment that we can tell if the image is a successful docker image by the successful running of the final application from Openshift. A successful docker image (see Figure 9 ), should contain all the necessary library packages and the inspection part, including the business post-processing logic when the image passes through the dispatcher API.

\paragraph{Log4j security concerns }
It is also worth paying attention to the security of Jenkins-based semi-automatic deployment methods, such as log4j security concerns caused by Java \cite{log4j}.

\subsection{Run In Fly: OpenShift VS Kubernetes}
OpenShift is a cloud development Platform as a Service (PaaS) developed by Red Hat, enabling developers to develop and deploy their applications on cloud infrastructure \cite{openshiftonline}. It is a well-established platform and relatively straightforward for AI engineers to get started and deploy an application based on its learning cost compared to other PaaS platforms (see in Table \ref{comparsion}).  There are many benefits adopting Openshift-based PaaS services in real industrial applications, such as graphical operations. AI developers can get on board quickly, even if unfamiliar with the backend architecture since backend support is available from this platform as soon as the docker image is uploaded successfully.

\paragraph{OOM killed: resource concerns}
Besides, this approach has also many shortcomings. It is often the case that after deployment, due to resource issues, OOM (Out of memory killed), frequently occurs for a single use case (see in Figure 7. Poor scalability is also a critical issue when resources become very expensive while simultaneously needing to process a massive amount of data. Limited storage and CPU resources of OpenShift often make people give up on this platform.

\paragraph{Transparency concerns}
The transparency is a topic of great concern to computer vision algorithm engineers as well. Engineers often need to index logs, while Openshift often has only a limited number of inspection logs and often connects to Kibana \cite{kibana} ; it takes much time to query these columns. Meanwhile, OpenShift reserved many useless logs for network layer ping pang, which accelerated resource consumption while increasing costs for the enterprise.

\subsection{Cloud Migration bring new CI/CD Pipeline}
To retain the benefits of OpenShift deployments, it may be a good choice to go with a more cost-effective cloud provider, but cloud migration also brings a lot of new tools and challenges, such as K9s, or the ArgoCD-based deployment tools that follow \cite{ArgoCD}. More obviously, this approach brings more testing in terms of docker image.  This feature allows AI engineers to realistically check the success of a docker image build. \cite{kibana}

\subsection{How ArgoCD leverage the Deployment}
 
As shown in the Figure 8, ArgoCD is also a Kubernetes-based deep learning application deployment tool and controller. Similar to Jenkins, they both have the same automation feature where any changes made to the desired target state in the Git repository can be automatically applied and reflected in the specified target environment. Unlike PaaS-based deployments such as Openshift, here we can monitor the server's state in real-time and easily modify or refresh, delete, and synchronize the state of Pods. With repository servers, many operations involve commits, tags, and branches, so a higher level of git knowledge is required. In addition, it is straightforward to implement local testing of docker images. Logs and CPU resources for individual pods are no longer an issue. Once the model is deployed successfully, it can run stably and will not encounter OOM problems like the Openshift platform-based deployment approach \cite{ArgoCD}.

\paragraph{Parallelism concerns}
Similarly, server and git-based deployment methods have many issues; for example, one of the highlights of this deployment method is the need for the tag, which requires many git operations, so if the team is deploying multiple use cases at the same time, parallelism will become very problematic, which requires a dedicated person to control on git merge operations. The consequence is that each deployment takes twice as long as the previous Openshift-based methods. Since all CPU parameters of the model are controlled by some configuration, each modification of these parameters requires a new git operation, so it becomes very cumbersome for the rescaling and redeployment process.

\subsection{ Automatic Deployment based on GitHub, Docker and ArgoCD }

Whether based on Openshift or ArgoCD, the model deployment process requires more or less manual effort to build the docker image through a series of command lines. Assuming a wrong step on the command line, or a wrong version number of a python package, these will directly lead to the failure of the docker image, and the process of building docker requires the installation of all the environment files and packages, which is no less than installing a single software. General principles dictate that the docker image is the sandbox that allows the AI model to run successfully. How to automate or semi-automate this process can be done using ArgoCD, Helm, and Git. The diagram below Figure 10 gives a decent overview of this process.

With the above approach, we automate the generation of docker images, but this can also bring up issues of low transparency and difficulty in local testing.

\section{Challenges and Evaluation Metrics}
\label{gen_inst}
\textbf{Learning cost for engineers}: This is directly related to the educated-market of this deployment method. If it is a very mature company with operating system support like PaaS, it may be speedy for deployment and easy to learn, which is necessary for enterprises to build their initial machine learning system.

\textbf{Transparency}: of a deployment technique or deployment system will directly determine the speed of debugging and later find the cause of model problems. Academia is very enthusiastic about the accuracy of artificial intelligence, but less popular papers can be seen on the interpretability and stability of models \cite{datamining2}. In industrial applications, we often encounter situations where models perform very well in training and testing, but face problems once deployed - so transparency in all of this should be important, starting with model training, dataset selection, business code for edge cases, transparency of docker images and accessibility of deployment logs.

\textbf{Stability \& Security}: for example, when the model is deployed successfully, the model should be able to run relatively stable. Here, particular emphasis is placed on the Stability of the machine learning system deployment pipeline. Different deployment methods have many advantages and disadvantages, from static to semi-automatic and from static to automatic (see in table \ref{comparsion}). Moreover, if the enterprise often changes the deployment method for reasons such as saving economic costs, it will cause the consequences of system instability. Because machine learning engineers need to spend time and effort to test and learn new deployment methods and migration, this is not conducive to improving model stability, which can also directly lead to delays in product delivery.

\textbf{Parallelism}: for machine learning deployment, large teams often have very high requirements for parallel deployment, where models, docker images, and deployment files should remain independent and uncorrelated \cite{docker}. High independence can ensure that the source of AI problems can be found quickly during the debugging process, and high Parallelism can ensure the rapid deployment of AI models and, thus, the rapid delivery of products.

\textbf{Cost}: the choice of deployment method often depends on the cost of this deployment method, which is one of the most important reasons for the enterprise to choose this method and the core reason for going through many iterations. However, for PaaS-based deployment methods such as OpenShift \cite{openshiftonline}\cite{Redhat}, the price of different cloud platforms should not be the ultimate measure. The wisest approach should be to choose the most suitable solution after measuring the scale of the current machine learning system; otherwise, frequent changes on deployment methods or cloud platform providers in pursuit of the cheapest solution can also lead to unstable machine learning systems.

\section{Conclusion}
\label{others}

This paper discusses the advantages and disadvantages of different deployment methods and the challenges encountered from the perspective of the practical large-scale industrial machine learning system. We propose the metrics and strategies that should also be measured when choosing a deployment method. In this paper, we also discussed some current misconceptions and common mistakes made from the perspective of engineers and stakeholders. This paper can be a good reference for choosing a suitable deployment method for Production ML Systems and a handbook for debugging reference schemes.

\appendix

\section{Appendix}

\begin{table}[b]
\caption{General workload for machine learning System Development. As stated from \cite{datamining2}, 80 percent of modern ML project delivery time are spend on pre-evaluation and model understanding and training occupies 20 percent. However, the right strategy for deployment and techniques are lessly discussed \cite{datamining2}. }
\label{workload}
 \resizebox{\columnwidth}{14mm}{ 
\begin{tabular}{l l l l l l}\toprule

\textbf{Business Understanding} & \textbf{Data Understanding} & \textbf{Data Preparation} & \textbf{Modelling} & \textbf{Evaluation} & \textbf{Deployment} \\[0.5ex]
\hline\\ 
Determine Business & Collect Initial Data & Select Data & Select Modeling Techniques & Evaluate Results& Plan Deployment 
\\[0.5ex]
 
Assess Situation & Describe Data & Clean Data & Generate Test Design & Review Process & Plan Monitoring and Maintenance  \\[0.5ex]

Determine Data Mining Goals  & Explore Data & Construct Data & Build Model & Determine Next Steps & Produce Final Report \\[0.5ex]

Produce Project Plan & Verify Data Quality & Integrate Data & Assess Model &  & Review Project \\[0.5ex]

 &  &  Format Data &  &  &    \\
 &  &  Dataset &  &  &    \\
\bottomrule\end{tabular}}
\end{table}

\begin{figure}[h]
  \centering
  \label{evaluationmetric}
  \includegraphics[width=0.6\textwidth]{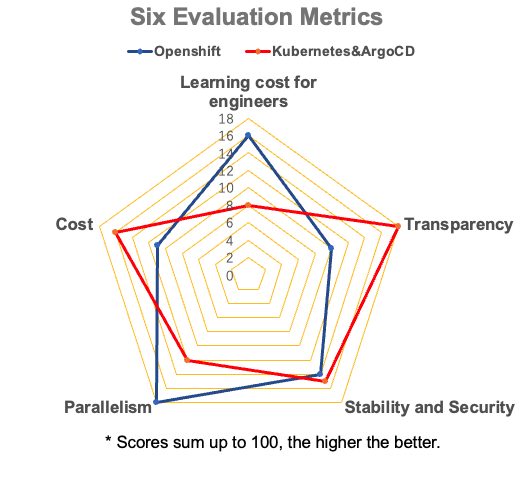}
  \includegraphics[width=1.0\textwidth]{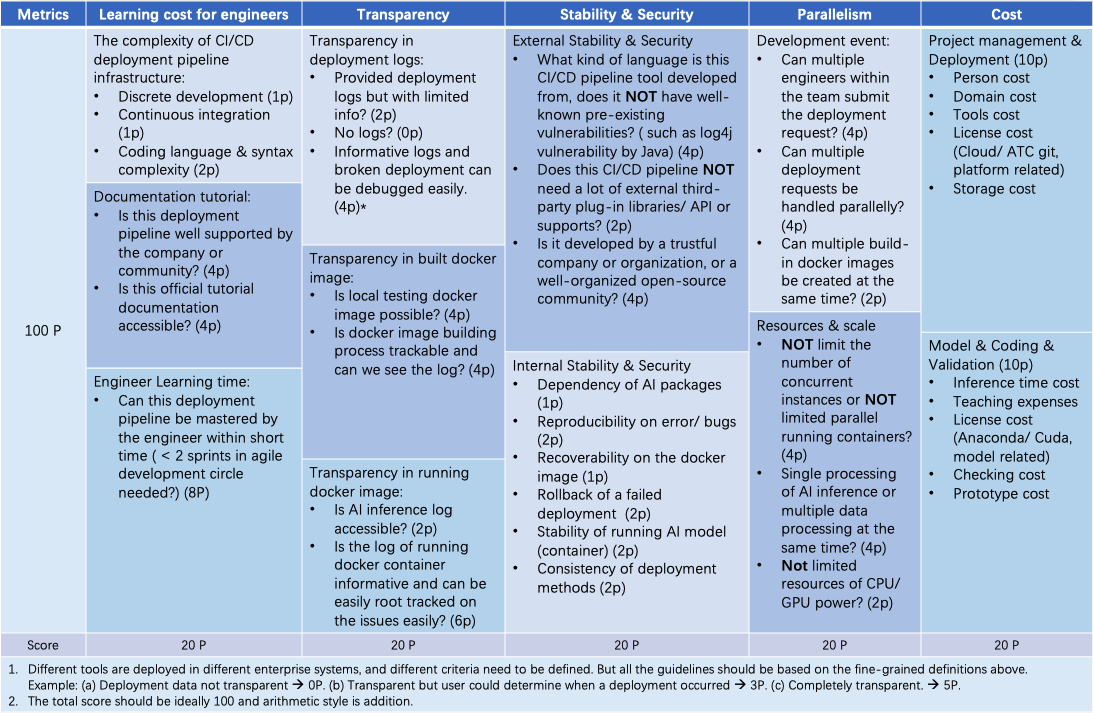}
  \caption{Our Six Evaluation Metrics. Final score can be calculated from Six Metrics (Learning cost for engineers, Transparency, Stability, Security, Parallelism and Cost) and each standard could be calculated by using the referred points. The standards are derived from Engineer experiences and summaries. The final conclusion could be based on real industry needs and could be weight differently according to real scenarios, as displayed visually in the score distribution graph (e.g. blue line for Openshift and red for Kubernetes\& ArgoCD as example). When the score is 100, it will appear as a regular pentagon.}
\end{figure}

\begin{table}[b]
\caption{ Sample comparison between OpenShift and Kubernetes based on Six Evaluation Metrics  \cite{openshiftonline}\cite{kubernetes}.}
\label{comparsion}
\resizebox{\columnwidth}{!}{%
\begin{tabular}{@{}cl@{}}
\toprule
\textbf{Metrics}            & \multicolumn{1}{c}{\textbf{OpenShift}}                                          \\ \midrule
Learning cost for engineers & Openshift is a product from IBM running                                         \\
                            & on top of Kubernetes with other CI/CD integration like Jenkins and Ansible \cite{openshiftonline}      \\ \cmidrule(l){2-2} 
Transparency                & Image Streams and Service Catalog make OpenShift better in terms of management,   \\
                            & while limited logs also prohibiting advanced transparency \cite{openshiftonline}                      \\ \cmidrule(l){2-2} 
Stability and Security      & Often occurs Out of Memory Killed error, very strict in security                \\
                            & and need to certain level of permission to maintain minimum security level      \\ \cmidrule(l){2-2} 
Parallelism                 & Very high level of parallelism, OpenShift provides better support to users,     \\
                            & also enable of parallel deployment activities                                   \\ \cmidrule(l){2-2} 
Cost                        & Price is different for individual developer or entrepreneur,                    \\
                            & self-developer is free, but expensive for entrepreneur in terms of resources  \cite{openshiftonline}  \\ \midrule
\textbf{Metrics}            & \multicolumn{1}{c}{\textbf{Kubernetes}}                                         \\ \midrule
Learning cost for engineers & Kubernetes is an open-source container                                          \\
                            & orchestration tool fitting for industry project \cite{kubernetes}                                 \\ \cmidrule(l){2-2} 
Transparency                & Management of container images is not so easy in Kubernetes,                    \\
                            & also Kubernetes service catalog has less provision of services within clusters, \\
                            & however with help of ArgoCD or K9s can easily access to logs of deployment \cite{ArgoCD}      \\ \cmidrule(l){2-2} 
Stability and Security      & Very stable once container or pod is successfully deployed,                     \\
                            & easy to maintain security level                                                 \\ \cmidrule(l){2-2} 
Parallelism                 & Quiet limited, often requires additional tools for good deployment experience   \\ \cmidrule(l){2-2} 
Cost                        & Open-source tool, often not as single solution,                                 \\
\multicolumn{1}{l}{}        & needs third part plugins and cloud providers  \cite{kubernetes}                                   \\ \bottomrule
\end{tabular}%
}
\end{table}

\begin{figure}[h]
  \centering
  \label{fig:NSSM}
  \includegraphics[height=0.6\columnwidth]{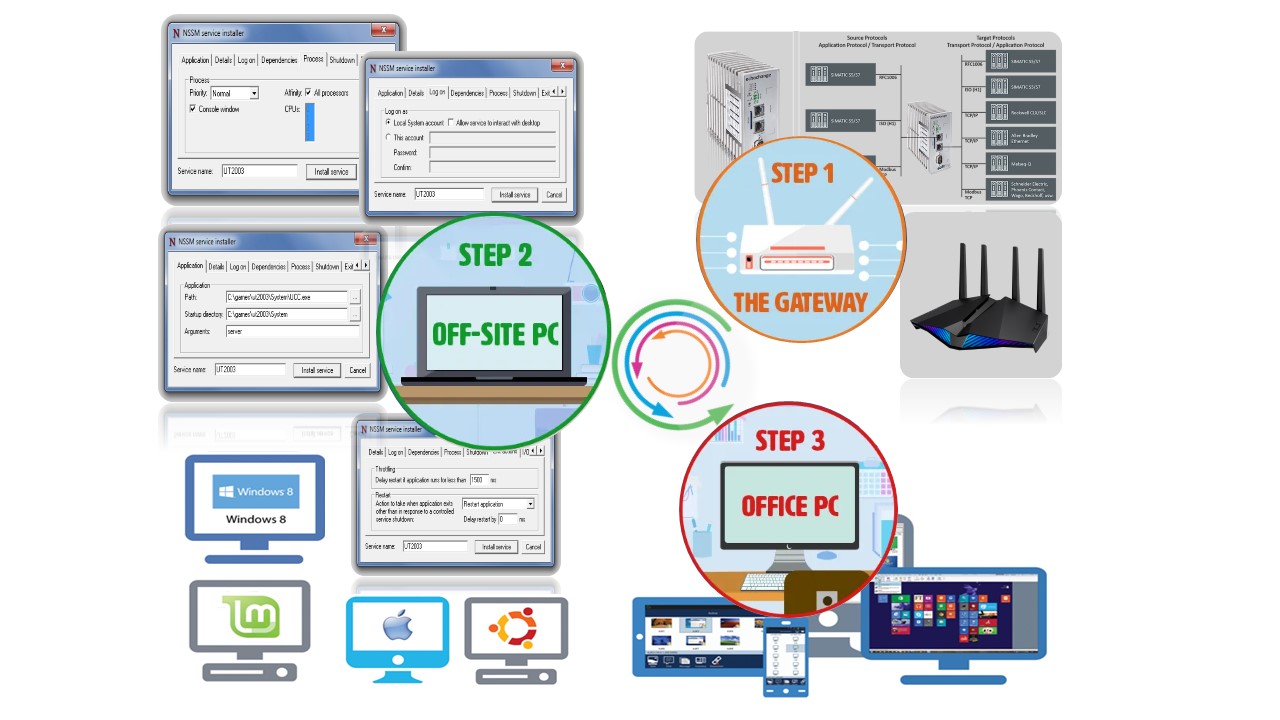} 
  \caption{NSSM Batch Service Pipeline. The Net-support Manager and NSSM Batch are usually utilized here for static deployment. The first step needs to purchase the Network Gateway system, such as IoT Gateway or industrial Network Gateway. The gateway usually contains not only the hardware device but also network protocols. then, it is about the set up of the NSSM service to run the inference script once off sit PC is started using local CPU/GPU resources. Moreover, Office PC can finally access off-site PC via Net-support Manager, which supports almost all 20+ systems. AI developers can finish the remote deployment of ML systems even without the Knowledge of real production environments.\cite{NSSM}}
\end{figure}

\begin{figure}
  \centering
  \label{Batchservice}
  \includegraphics[height=0.4\columnwidth]{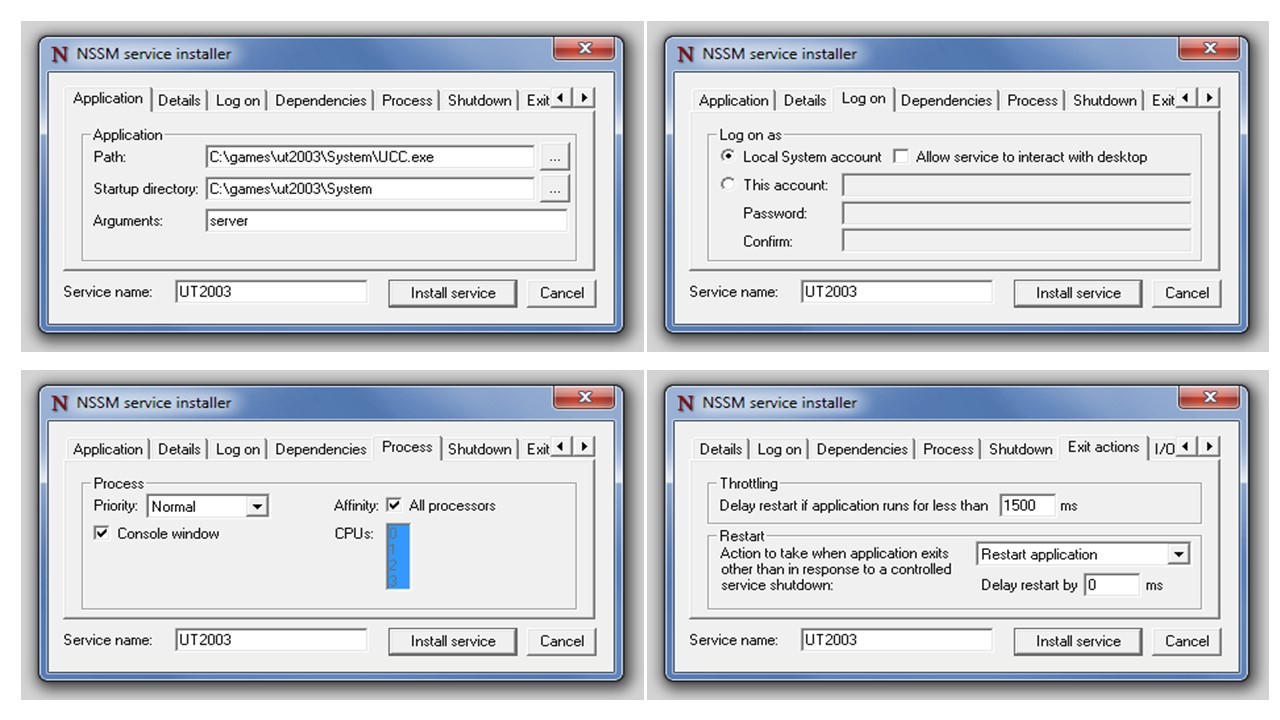} 
  \caption{Batch Service configuration \cite{NSSM}. These four figures summarize the setup of the Batch Service configuration. The top left illustrates the application path and start-up directory, e.g., it could be a python inference script or executable software path. The top right indicates the service logs and can be placed under the same project folder path. We can set the process-related settings such as CPU resources, as shown in the bottom left, and service restarting times, as captured in the bottom right. It is evident that NSSM services have many steps of configurations set, more steps or parameters as documented in following Figure 5.}
\end{figure}

\begin{figure}[h]
  \centering
  \label{NSSMcommand}
  \includegraphics[width=1\columnwidth]{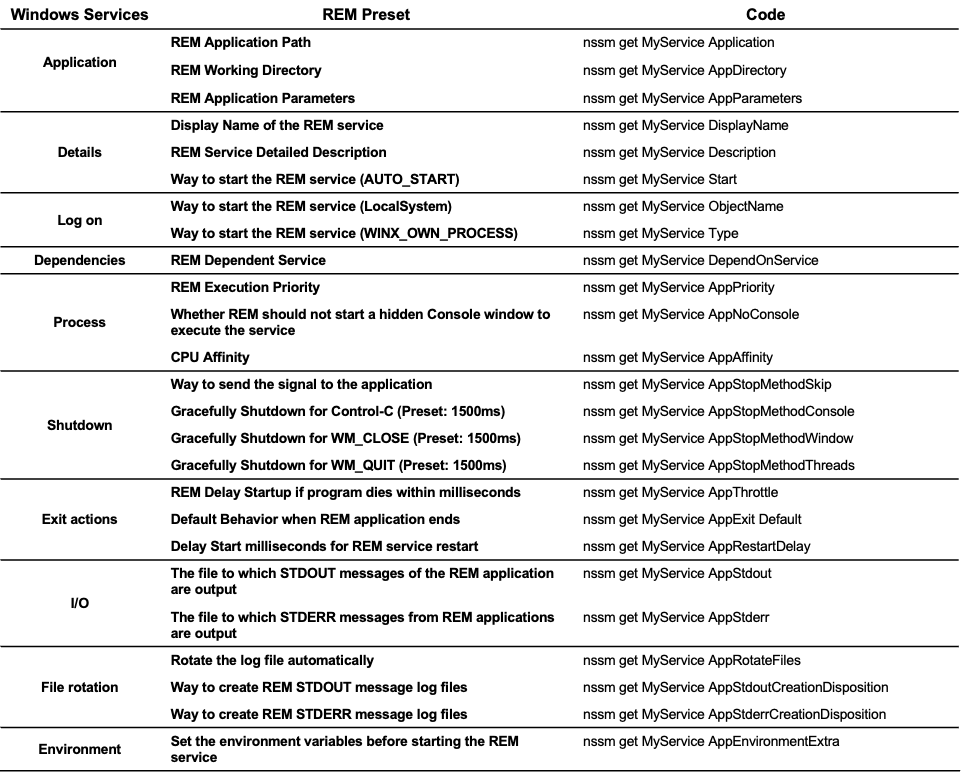} 
  \caption{NSSM Command Window Description \cite{NSSM}. Equivalent command examples upper show the commands which would configure an existing service to NSSM Command Window. If the application needs to start in a particular directory you can enter it in the Startup directory field. If the field is left blank the default startup directory will be the directory containing the application. Initial parameters are often default parameters.}
\end{figure}

\begin{figure}[h]
  \centering
  \includegraphics[width=1\columnwidth]{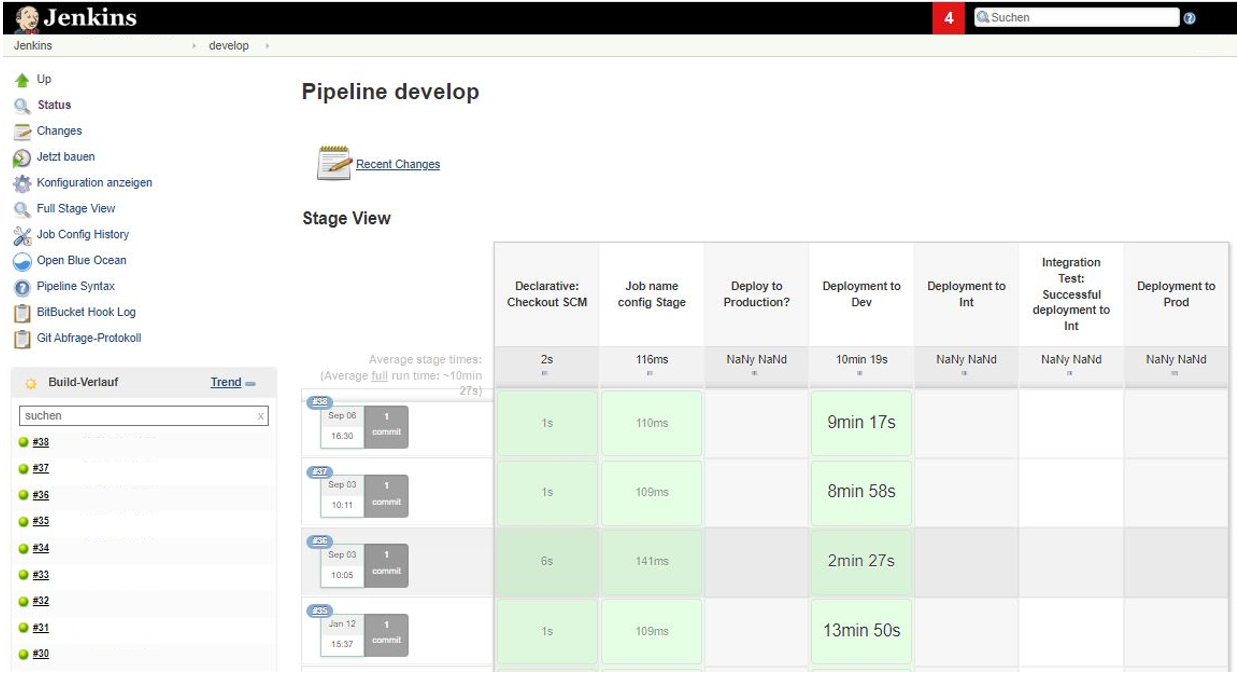} 
  \label{JenkinsUI}
  \caption{Jenkins Deployment User Interface (Image source from \cite{Jenkins}). As an open-source CI automation server, Jenkins can be used to automate all sorts of tasks related to building, testing, and delivering or deploying \cite{Jenkins}. Jenkins can invoke docker containers on agents/nodes (from a Jenkinsfile or through Blue Ocean) to build AI Pipeline projects, namely, write a Jenkins script (Jenkinsfile is recommended) and clearly define each stage and step, then add credential into the requirment.txt file. As shown in the figure, from Graphic Jenkins Pipeline, we can easily trigger the docker image build and track the stage of current deployment in real-time \cite{Jenkins}.}
\end{figure}

\begin{figure}[h]
  \centering
  \label{OS Memeory}
  \includegraphics[width=1\columnwidth]{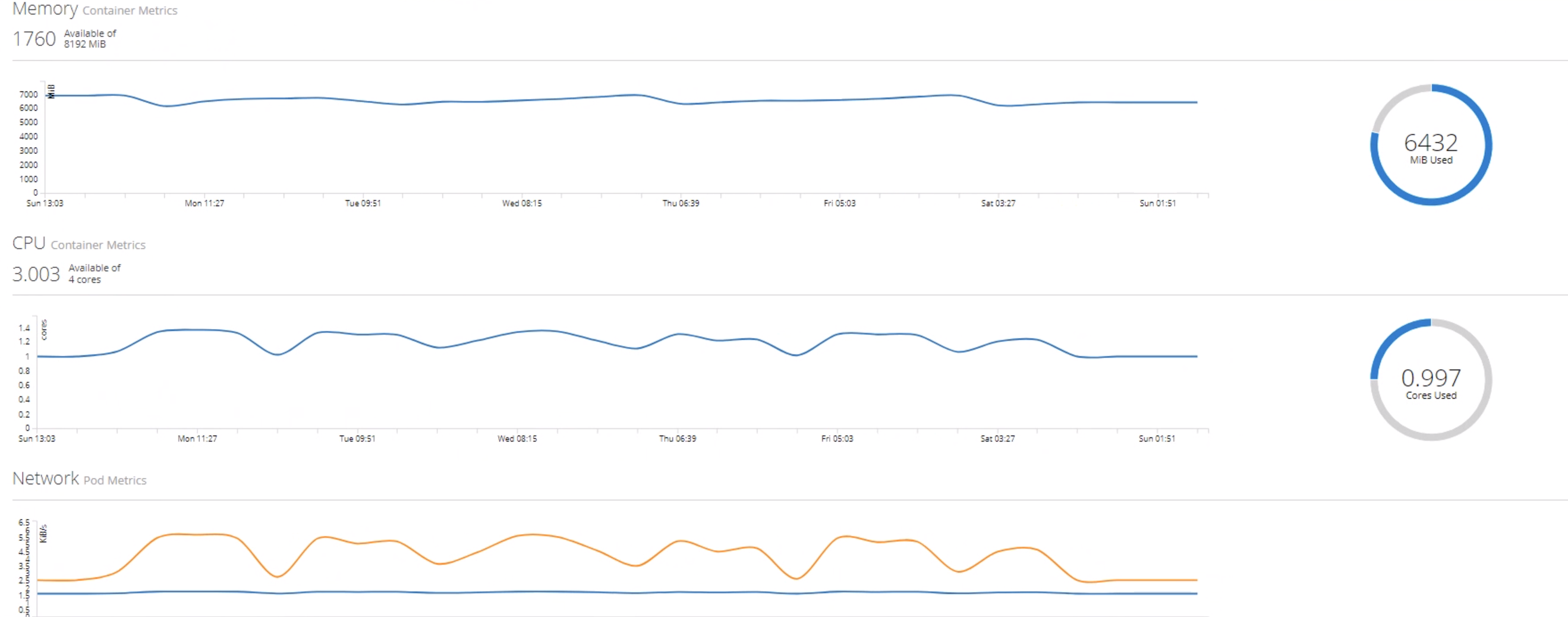} 
   
  \caption{Use case resource consumption logs based on PaaS OpenShift deployment methods \cite{openshiftonline}. These three logs perfectly illustrate the consumption of a typical anomaly detection use case running on Openshift. Use-case deployed through this method often encounters the OOM ( Out of Memory Killed) error, while the docker image size is usually up to 3 to 4 GiB. Moreover, the image size is up to 4MB. The setting for 4 to 6 CPU cores limit sometimes is also not enough. From practical experiments and experiences, the PyTorch weight is much less than TensorFlow save model weight. Although the memory composition also depends on the post-processing logic complexity and model amount, frequent OOM Killed issues while saving many useless network layer logs make people less trusted in the stability of the AI systems.}
\end{figure}

\begin{figure}[h]
  \centering
  \label{ArgoCDCICD}
  \includegraphics[width=1\columnwidth]{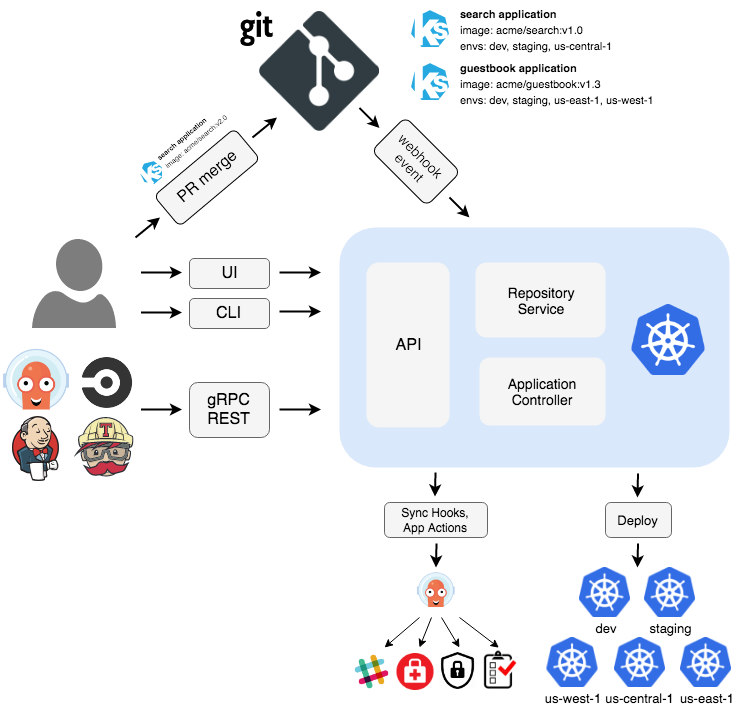} 
  \caption{ArgoCD CI/CD pipeline \cite{ArgoCD}. The Graph summarizes the structure of the Repository container and Kubernetes-based deployment in great detail. From left to right, tools like ArgoCD and Jenkins continuously monitor all running applications and serve as Kubernetes controllers. When the AI developer builds the docker image and pushes the docker image to the cloud repository, the only thing left is to trigger the pull request. ArgoCD will compare their live state to the desired state specified in the Git repository; once it is differentiated, it will automatically start pulling the latest docker files and all necessary configuration files and start the SYN and Deploy. Here, Kubernetes is responsible for the Deployment and Management of infrastructure. For a detained explanation of docker image, you may refer to following Figure 9.}
\end{figure}

\begin{figure}[h]
  \centering
  \label{docker}
  \includegraphics[width=1\columnwidth]{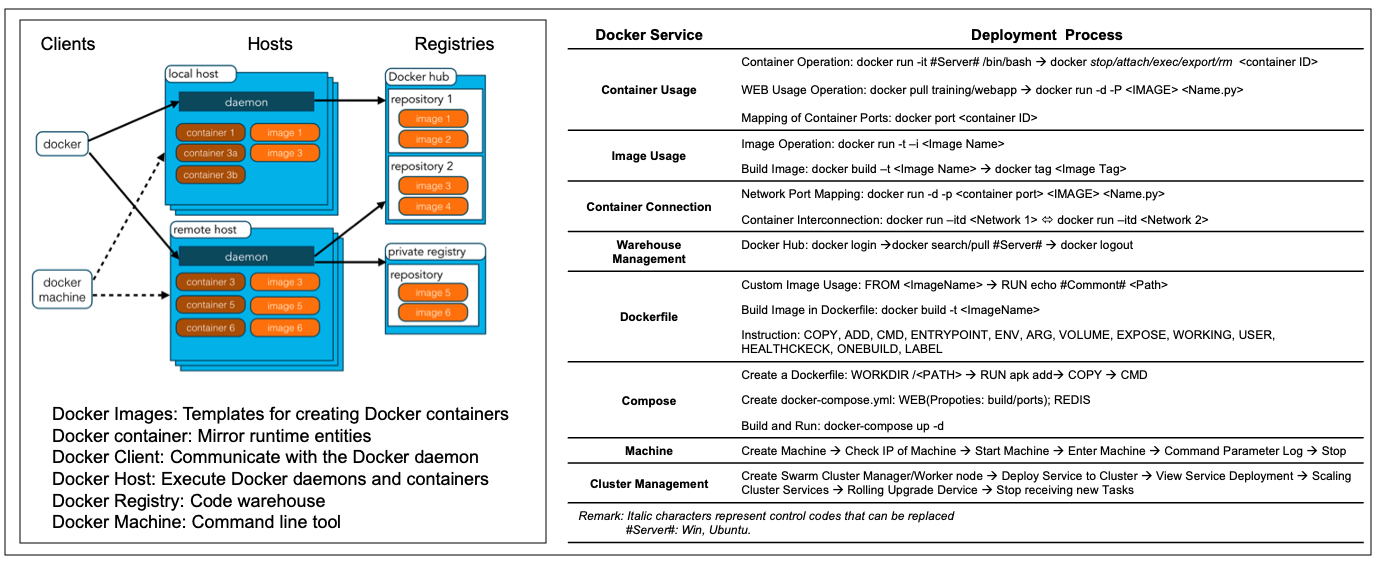} 
  \caption{Docker architecture. Docker image is equivalent to a root file system. The relationship between Image and Container is like classes and instances in object-oriented programming \cite{docker}. Image is a static definition, and a container is an entity when the image is run. Containers can be created, started, stopped, deleted, suspended, etc. Repository can be regarded as a code control center to store images. The rest of the units are well explained in the graph upper. }
\end{figure}

\begin{figure}[h]
  \centering
  \label{ArgoCD Web UI}
  \includegraphics[width=1\columnwidth]{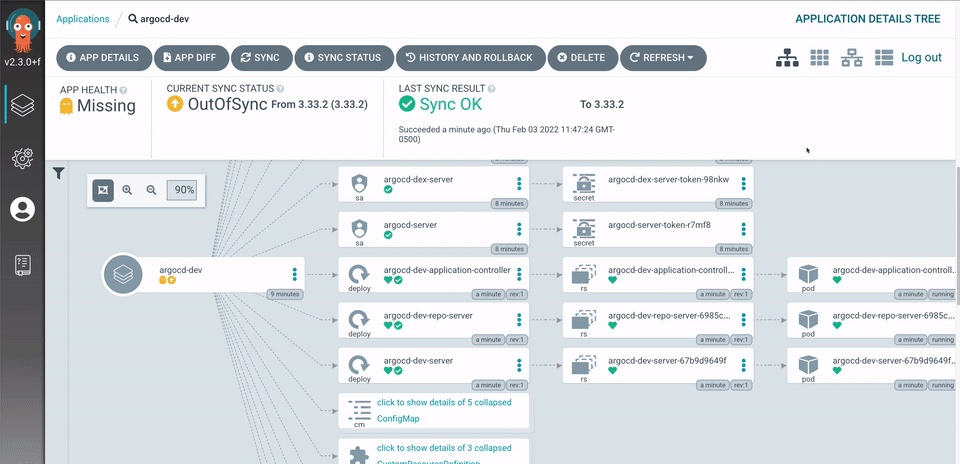} 
   
  \caption{ArgoCD Application User Interface (Image source from \cite{kubernetes}). As a declarative, GitOps continuous delivery tool for Kubernetes \cite{kubernetes}, ArgoCD can track the application status in Kubernetes Repository \cite{ArgoCD}. and automates the deployment of the desired application states to specified target environments. It can update branches, tags, or pins to a specific version of manifests with a Git commit\cite{ArgoCD}. The graph shows that the Web UI provides a real-time view of application activity. It supports multiple config management/templating tools (Customize, Helm, Jsonnet, plain-YAML)\cite{ArgoCD}. While it also requires permission to perform centrin level of delete/refresh/edit/SYN) actions as well.}
\end{figure}

\end{document}